\pdfoutput=1
\documentclass[11pt]{article}

\usepackage[]{ACL2023}

\usepackage{times}
\usepackage{latexsym}

\usepackage[T1]{fontenc}
\usepackage[utf8]{inputenc}

\usepackage{microtype}

\usepackage{amsmath,amsfonts,bm}

\def\eqref#1{equation~\ref{#1}}
\def\1{\bm{1}}

\def\vy{{\bm{y}}}

\DeclareMathAlphabet{\mathsfit}{\encodingdefault}{\sfdefault}{m}{sl}
\SetMathAlphabet{\mathsfit}{bold}{\encodingdefault}{\sfdefault}{bx}{n}

\makeatletter

\newcommand*\totht[1]{\dimexpr\ht#1+\dp#1\relax}
\newcommand*\leading{{\setbox0\hbox{\strut}\the\totht0}}
\newcommand*\fntsize{{\setbox0\hbox{Mg}\the\totht0}}% total size of the letters Mg
\newcommand*\showsize[1]{{#1 {\ttfamily\string#1} (\f@size pt) \fntsize/\leading}\par}
\makeatother

\title{DAMO-NLP at SemEval-2023 Task 2: A Unified Retrieval-augmented System for Multilingual Named Entity Recognition}
\usepackage{microtype}
\usepackage{bbm}
\usepackage{amsmath}
\usepackage{amsfonts}
\usepackage{algorithm}
\usepackage{algpseudocode}
\usepackage{booktabs}
\usepackage{xparse}
\usepackage{graphicx}
\usepackage{multicol}
\usepackage{multirow}
\usepackage{tikz}
\usetikzlibrary{shapes,arrows}
\usepackage{natbib}
\usepackage{caption}
\usepackage{subcaption}
\usepackage{tikz-dependency}
\usepackage{wrapfig}
\usepackage{pgfplots}
\pgfplotsset{compat=1.17} 
\usepackage[T1]{fontenc}
\usepackage{subfiles}
\usepackage{readarray}
\usepackage{xcolor}
\usepackage{import}
\usepackage{hhline}
\usepackage{enumitem}
\usepackage{boldline}
\usepackage{mathtools}
\usepackage{colortbl}
\usepackage{cleveref}
\usepackage{fontawesome}

\crefname{section}{§}{§§}
\Crefname{section}{§}{§§}

\usepackage{pifont}% http://ctan.org/pkg/pifont
\newcommand{\cmark}{\ding{51}}%
\newcommand{\xmark}{\ding{55}}%
\definecolor{bg}{rgb}{0.95, 0.95, 0.95}

\newcommand{\RNum}[1]{\uppercase\expandafter{\romannumeral #1\relax}}

\newcommand{\shtu}{\textsuperscript{\faSunO}}
\newcommand{\zju}{\textsuperscript{\faMoonO}}
\newcommand{\damo}{\textsuperscript{\faStarO}}
\newcommand{\thu}{\textsuperscript{\faHeartO}}

\newcommand{\tabincell}[2]{\begin{tabular}{@{}#1@{}}#2\end{tabular}}

\author{Zeqi Tan\zju$^{\dagger}$, 
Shen Huang\damo$^{\dagger}$, 
Zixia Jia\shtu$^{\dagger}$, 
Jiong Cai\shtu$^{\dagger}$, 
Yinghui Li\thu$^{\dagger}$, 
Weiming Lu\zju \\
\textbf{
Yueting Zhuang\zju, 
Kewei Tu\shtu, 
Pengjun Xie\damo, 
Fei Huang\damo, 
Yong Jiang\damo\thanks{\;\,: project lead. $^\dagger$: equal contributions. This work was done during Zeqi Tan, Zixia Jia, Jiong Cai, and Yinghui Li’s internship
at DAMO Academy, Alibaba Group.}} \\
\damo DAMO Academy, Alibaba Group \\
\zju College of Computer Science and Technology, Zhejiang University \\
\shtu School of Information Science and Technology, ShanghaiTech University \\
\thu Tsinghua Shenzhen International Graduate School, Tsinghua University \\
{\tt \{zqtan,yzhuang,luwm\}@zju.edu.cn} ~ {\tt liyinghu20@mails.tsinghua.edu.cn}\\
{\tt \{jiazx,caijiong,tukw\}@shanghaitech.edu.cn}\\
{\tt \{pangda,chengchen.xpj,f.huang,yongjiang.jy\}@alibaba-inc.com}}

\begin{document}
\maketitle
\begin{abstract}
% The MultiCoNER \RNum{2} shared task aims to tackle multilingual named entity recognition (NER) in fine-grained and noisy scenarios, and it inherits the semantic ambiguity and low-context setting of the MultiCoNER \RNum{1} task.
% To cope with these problems, the previous top systems in the MultiCoNER \RNum{1} either incorporate the knowledge bases or gazetteers.
% However, they still suffer from insufficient knowledge, limited context length, single retrieval strategy.
% In this paper, our team \textbf{DAMO-NLP} proposes a unified retrieval-augmented system (U-RaNER) for fine-grained multilingual NER.
% We perform error analysis on the previous top systems and reveal that their performance bottleneck lies in insufficient knowledge.
% Also, we discover that the limited context length causes the retrieval knowledge to be invisible to the model. To enhance the retrieval context, we incorporate the entity-centric Wikidata knowledge base, while utilizing the infusion approach to broaden the contextual scope of the model. 
% Also, we explore various search strategies and refine the quality of retrieval knowledge.
% Additionally, we perform this task using \textbf{ChatGPT}, which has received the most publicity recently, 
% and the results show that there is still much room for improvement on the extraction task.
% Our system wins 9 out of 13 tracks in the MultiCoNER \RNum{2} shared task. We will release the dataset, code, and scripts of our system at {\small \url{ https://github.com/modelscope/AdaSeq/tree/master/examples/U-RaNER}}.

The MultiCoNER \RNum{2} shared task aims to tackle multilingual named entity recognition (NER) in fine-grained and noisy scenarios, and it inherits the semantic ambiguity and low-context setting of the MultiCoNER \RNum{1} task. To cope with these problems, the previous top systems in the MultiCoNER \RNum{1} either incorporate the knowledge bases or gazetteers. However, they still suffer from insufficient knowledge, limited context length, single retrieval strategy. In this paper, our team \textbf{DAMO-NLP} proposes a unified retrieval-augmented system (U-RaNER) for fine-grained multilingual NER. We perform error analysis on the previous top systems and reveal that their performance bottleneck lies in insufficient knowledge. Also, we discover that the limited context length causes the retrieval knowledge to be invisible to the model. To enhance the retrieval context, we incorporate the entity-centric Wikidata knowledge base, while utilizing the infusion approach to broaden the contextual scope of the model. Also, we explore various search strategies and refine the quality of retrieval knowledge. Our system\footnote{We will release the dataset, code, and scripts of our system at {\small \url{https://github.com/modelscope/AdaSeq/tree/master/examples/U-RaNER}}.} wins 9 out of 13 tracks in the MultiCoNER \RNum{2} shared task. 
Additionally, we compared our system with ChatGPT, one of the large language models which have unlocked strong capabilities on many tasks. The results show that there is still much room for improvement for ChatGPT on the extraction task.

\end{abstract}

\section{Introduction}
\label{intro}

The MultiCoNER series shared task \citep{multiconer-report,multiconer2-report} aims to identify complex named entities (NE), such as titles of creative works, which do not possess the traditional characteristics of named entities, such as persons, locations, etc.
It is challenging to identify these ambiguous complex entities based on short contexts \citep{DBLP:journals/corr/AshwiniC14,meng2021gemnet,fetahu-etal-2022-dynamic}.
The MultiCoNER \RNum{1} task \citep{multiconer-report} focuses on the problem of semantic ambiguity and low context in multilingual named entity recognition (NER).
In addition, the MultiCoNER \RNum{2} task \citep{multiconer2-report} this year poses two major new challenges: (1) a fine-grained entity taxonomy with 6 coarse-grained categories (\texttt{Location}, \texttt{Creative Work}, \texttt{Group}, \texttt{Person}, \texttt{Product} and \texttt{Medical}) and 33 fine-grained categories, 
and (2) simulated errors added to the test set to make the task more realistic and difficult, like the presence of spelling mistakes. 
% and typos.

\begin{figure}[t!]
  \centering
  \includegraphics[width=\linewidth]{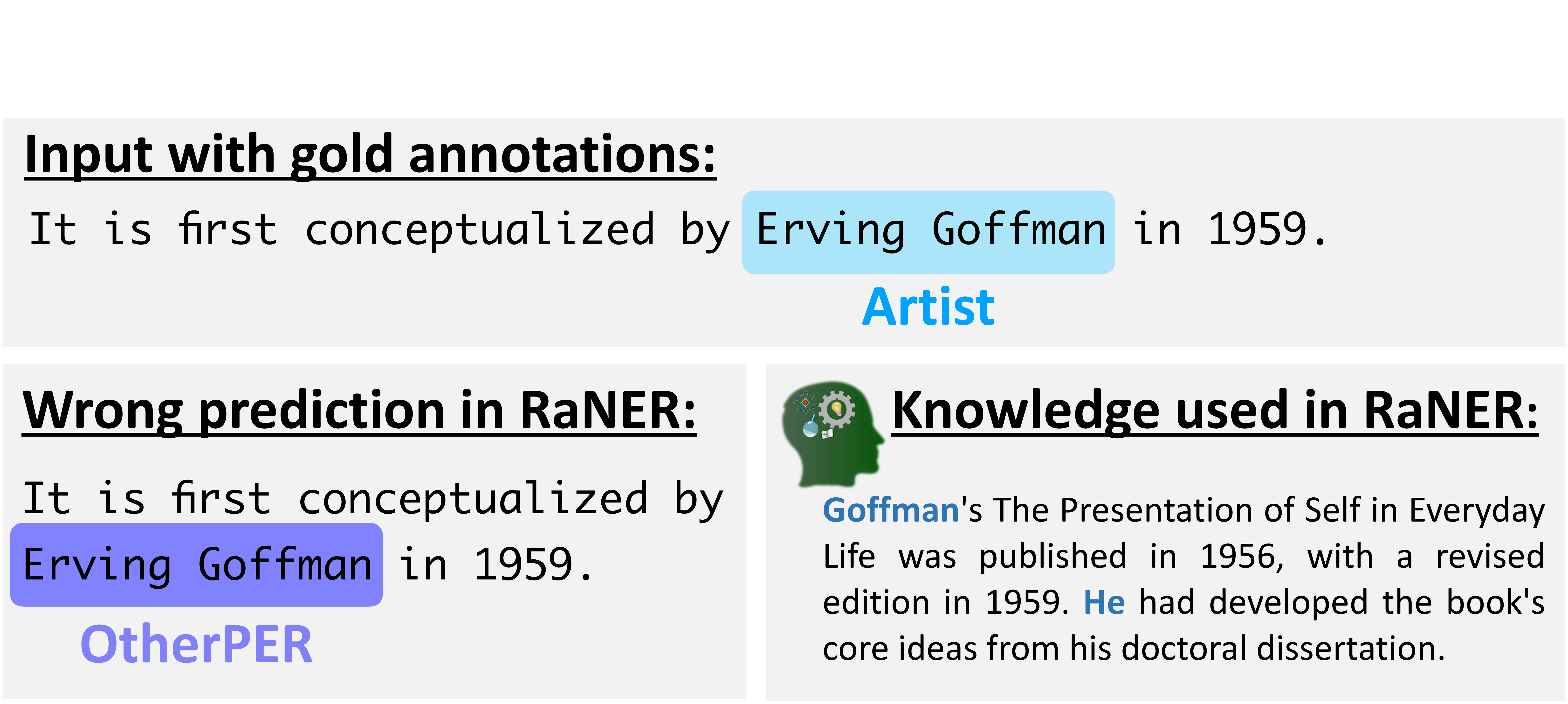}
  \caption{An example of wrong prediction in RaNER \cite{wang-etal-2022-damo}(one of the top systems  in the MultiCoNER \RNum{1} task \cite{multiconer-report}) . 
  % The colored underlines indicate the gold and the predicted entities respectively.
  % The green panel shows the external knowledge used by RaNER.
  This case illustrates that the knowledge covered is not sufficient for fine-grained complex NER.
}
  \label{fig:intro}
\end{figure}

The previous top systems \citep{wang-etal-2022-damo,chen-etal-2022-ustc} of the MultiCoNER \RNum{1} task incorporate additional knowledge in pre-trained language models, either a knowledge base or a gazetteer.
RaNER \citep{wang-etal-2022-damo} builds a multilingual knowledge base based on Wikipedia and the original input sentences are then augmented with retrieved contexts from the knowledge base, allowing the model to access more relevant knowledge.
GAIN \citep{chen-etal-2022-ustc} proposes a gazetteer-adapted integration network with a gazetteer built from Wikidata to improve the performance of language models.
Although these systems achieve impressive results, they still have some drawbacks.
\textbf{First}, insufficient knowledge is a common problem.
As shown in Figure \ref{fig:intro}, the knowledge used in RaNER can help the model to identify \textit{Erving Goffman} as a person, but cannot further determine the fine-grained category \texttt{Artist}.
% The dictionary utilized in \citet{chen-etal-2022-ustc} suffers from the absence of contextual information and is powerless against ambiguous entities.
\textbf{Second}, these methods mostly suffer from the limited context length. \citet{wang-etal-2022-damo} discards stitched text that is longer than 512 after tokenizing, which means that plenty of retrieved context is not visible to the model, leading to resource waste.
\textbf{Third}, these systems have a single retrieval strategy. \citet{wang-etal-2022-damo} acquires knowledge by text retrieval, while \citet{chen-etal-2022-ustc} accesses knowledge by dictionary matching. This single way of knowledge acquisition will result in the underutilization of knowledge.

% \paragraph{overview of the approach}
To tackle these problems, we propose a unified retrieval-augmented system (U-RaNER) for fine-grained multilingual NER.
We use both Wikipedia and Wikidata knowledge bases to build our retrieval module so that more diverse knowledge can be considered.
As shown in Figure \ref{fig:intro}, if we locate the entry for \textit{Erving Goffman} in Wikidata, we can make use of fine-grained entity category information to facilitate predictions.
Also, we discover that the retrieval context dropped by the model may also contain useful knowledge.
Thus, we explore the infusion approach to make more context visible to the model.
In addition, we use multiple retrieval strategies to obtain the most relevant knowledge from two knowledge bases, 
% so as to make the best use of the resources.
further improving the model performance.

Our main contributions are as follows:
\begin{enumerate}[leftmargin=*,noitemsep]
    \item We propose a unified retrieval-augmented system for fine-grained multilingual NER. Our system incorporates more diverse knowledge bases and significantly improves the system performance compared to baseline systems (Section \cref{sec:methods}, \cref{sec:exp})
    \item We initiated our investigation by identifying the primary bottleneck of the previous top-performing system, which we determined to be insufficient knowledge. Consequently, we focused on exploring both data and model enhancements to improve system performance. (Section \cref{sec:data})
    \item We employ multiple retrieval strategies to obtain entity information from Wikidata, in order to complement the missing entity knowledge. (Section \cref{sec:4.1})
    \item Additionally, we utilize the infusion approach to provide a more extensive contextual view to the model, thus enabling better utilization of the retrieved context (Section \cref{sec:4.2}).
    \item Extensive experimental analysis demonstrates the effectiveness of diverse knowledge sources and broader contextual scopes for improving model performance. (Section \cref{sec:exp})
\end{enumerate}

\section{Related Work}

Named Entity Recognition (NER) \cite{Sundheim1995NamedET} is a fundamental task in Natural Language Processing.
Because of the long-term attention and the rapid development of pre-trained language models, various models~\cite{akbik-etal-2018-contextual,devlin-etal-2019-bert,yamada-etal-2020-luke,wang-etal-2020-more,wang2020automated} have achieved state-of-the-art results and performance in general NER scenarios and datasets, such as CoNLL 2002~\cite{DBLP:conf/conll/Sang02}, CoNLL 2003~\cite{tjong-kim-sang-de-meulder-2003-introduction}, and OntoNotes 5.0~\cite{pradhan-etal-2013-towards}. 
Considering that the previous task settings or datasets are monolingual and scenario-constrained, the task of \textbf{Multi}lingual \textbf{Co}mplex \textbf{N}amed \textbf{E}ntity \textbf{R}ecognition (MultiCoNER) is proposed to promote the NER research to be more oriented to real scenarios~\cite{multiconer-report}. Our work focuses on this task and we will introduce the related work from the dataset and method of MultiCoNER respectively:

\paragraph{Challenges of MultiCoNER Dataset}  To address contemporary in the NER field, \citeauthor{multiconer-data} construct MultiCoNER, a large and complex dataset for Multilingual Complex Named Entity Recognition. This 26M token dataset covers 3 domains (including Wiki, question, and search query) and 11 languages (12 languages for SemEval-2023). In particular, aiming at the main challenges of NER research, the MultiCoNER dataset sets 4 key characteristics: 
(1) \textbf{Low Context}: Existing NER methods perform poorly if the context is less informative~\cite{meng2021gemnet}, thus, texts in MultiCoNER are low in context to assess the model's performance on the more realistic and difficult setting. 
(2) \textbf{Sufficient Diversity}: MultiCoNER contains an rich variety of entity types, both simple and difficult, which makes it possible to evaluate the model more comprehensively.
(3) \textbf{Reasonable Distribution}: Considering the non-negligible long-tail distribution problem faced by the previous datasets makes the construction of training data extremely difficult, MultiCoNER ensures that the distribution of its entities is more even and reasonable so that it can be evaluated comprehensively.
(4) \textbf{High Complexity}:  
Increasing the complexity of the dataset can effectively improve the quality of the dataset~\cite{fetahu2021gazetteer}. Therefore, in addition to monolingual subsets, MultiCoNER also distinctively contains a multilingual subset and a code-mixed one, which makes it more challenging. Note that in the dataset version of SemEval-2023, this challenge and setting do not exist.

% If you published a paper at the task last year, you should cite it here.
\paragraph{Progress of MultiCoNER Methods} With the MultiCoNER dataset as the core, the SemEval-2022 Task 11 attracts 236 participants, and 55 teams successfully submit their system~\cite{multiconer-report}. Among them, there are many successful and excellent works worthy of discussion. 
DAMO-NLP~\cite{wang-etal-2022-damo} proposes a knowledge-based method that gets multilingual knowledge from Wikipedia to provide informative context for the NER model. And they achieve the previous best overall performance on the MultiCoNER dataset.
USTC-NELSLIP~\cite{chen-etal-2022-ustc} proposes a gazetteer-adapted integration network to improve the model performance for recognizing complex entities.
QTrade AI~\cite{gan-etal-2022-qtrade} designs kinds of data augmentation strategies for the low-resource mixed-code NER task.
\emph{Previous efforts and studies on the MultiCoNER dataset have shown that external data and beneficial knowledge are essential to improve the performance of NER models on it}.

\paragraph{Retrieval-augmented NLP Methods} Retrieval-augmented techniques have proven to be highly effective in various natural language processing (NLP) tasks, as evidenced by the exceptional performance achieved in prior studies \cite{lewis2020retrieval,khandelwal2019generalization,borgeaud2022improving}. These approaches usually contain two parts: an information retrieval module and a task-specific module. Specifically, in the context of named entity recognition (NER), \citet{wang-etal-2021-improving} proposes leveraging off-the-shelf search engines like Google to retrieve external information and enhance the contextual representations of tokens in the input text, resulting in improved performance. Furthermore, subsequent research has focused on developing task-specific retrieval systems for domain-specific NER and multi-modal NER tasks, respectively \cite{zhang-etal-2022-domain,wang-etal-2022-named}. Drawing upon these insights, our proposed system is designed and optimized with guidance from these previous works.
% explain its relation with EntQA, RaG, KNN-LM, and so on.

\paragraph{Large Language Models On IE} 
Recent advances in NLP scale the parametric number of language models to hundreds of billions and have achieved phenomenal performance, such as GPT3 \citep{brown2020language}, OPT-175B \citep{zhang2022opt}, Flan-PaLM \citep{chung2022scaling}, LLaMA \citep{touvron2023llama} and ChatGPT\footnote{\url{https://openai.com/blog/chatgpt/}}.
In the field of information extraction (IE), ChatIE \citep{wei2023zeroshot} first uses ChatGPT for extraction, and the results show that there is still much room for improvement. 
Recent work experiments with instruction fine-tuning \citep{wang2023instructuie} and simplifying training objectives \citep{wang2023gptner} to adapt large language models to extraction tasks. 

\section{Data} \label{sec:data}

The MultiCoNER \RNum{2} corpus \citep{multiconer2-data} aims to recognize the complex named entities and pose new challenges for current NER systems. 
% With the same set of tags, the 12 multilingual datasets specifically include: BN-Bangla, DE-German, EN-English, ES-Spanish, FA-Farsi,  FR-French, HI-Hindi, IT-Italian, PT-Portuguese, SV-Swedish, UK-Ukrainian and ZH-Chinese.
To meet these challenges, we first reproduce the results of the top system \citep{wang-etal-2022-damo} and perform error analysis on validation sets.
We observe that the performance bottleneck of the system lies in the lack of knowledge.
Then, we investigate to break this bottleneck from data and model perspectives and improve model robustness.

Following \citet{wang-etal-2022-damo}, we build a multilingual KB based on Wikipedia of the 12 languages to search for the related documents.
We download the latest (2022.10.21) version of the Wikipedia dump from Wikimedia\footnote{\url{https://dumps.wikimedia.org/}} and convert it to plain texts.
We execute the official system on MultiCoNER \RNum{2} corpus and categorize the results according to whether the annotated entity appears in the retrieval context or not.
As shown in Table \ref{tab:entity}, the F1-measure on different types of test data differs significantly, e.g., 6.33\% on \texttt{DE} and 4.97\% on \texttt{ZH}.
This indicates that the lack of knowledge about entities in the retrieval context can have a significant impact on the model performance. 
With this insight, we consider data and model dimensions to compensate for this lack of knowledge.

\begin{table}[t!]
    \centering
    \small
    \scalebox{0.85}{
    \renewcommand\arraystretch{1.0}
    \begin{tabular}{llcccc}
    \toprule
        \textbf{Language} & \textbf{Data Type} & P & R & F1 & Ratio \\ 
    \midrule
        \multirow{4}[0]{*}{\textbf{BN}} & Total & 90.99  & 92.60  & 91.79  & 1.00  \\ 
         & In-context & 92.86  & 94.66  & 93.75  & 0.69  \\ 
         & Out-of-context & 88.06  & 89.39  & 88.72  & 0.31  \\ 
         % & \rowcolor[HTML]{EEF5FF} $\Delta$ & 4.80  & 5.27  & 5.03  & - \\ 
         & $\Delta$ & 4.80  & 5.27  & 5.03  & - \\ 
    \midrule
        \multirow{4}[0]{*}{\textbf{DE}} & Total & 81.83  & 83.00  & 82.41  & 1.00  \\ 
         & In-context & 83.80  & 88.11  & 85.90  & 0.54  \\ 
         & Out-of-context & 80.17  & 78.98  & 79.57  & 0.46  \\ 
         % & \rowcolor[HTML]{EEF5FF} $\Delta$ & 3.63  & 9.13  & 6.33  & - \\ 
         & $\Delta$ & 3.63  & 9.13  & 6.33  & - \\ 
    \midrule
        \multirow{4}[0]{*}{\textbf{ZH}} & Total & 76.71  & 78.40  & 77.54  & 1.00  \\
         & In-context & 79.27  & 83.87  & 81.50  & 0.26  \\ 
         & Out-of-context & 76.04  & 77.02  & 76.53  & 0.74 \\ 
         % & \rowcolor[HTML]{EEF5FF} $\Delta$ & 3.23  & 6.85  & 4.97  & - \\ 
         & $\Delta$ & 3.23  & 6.85  & 4.97  & - \\ 
    \bottomrule
    \end{tabular}
    }
\caption{The performance and ratio for different types of data on \texttt{BN}, \texttt{DE} and \texttt{ZH}.}
\label{tab:entity}
\end{table}

\begin{figure}[ht!]
  \centering
  \includegraphics[width=0.9\linewidth]{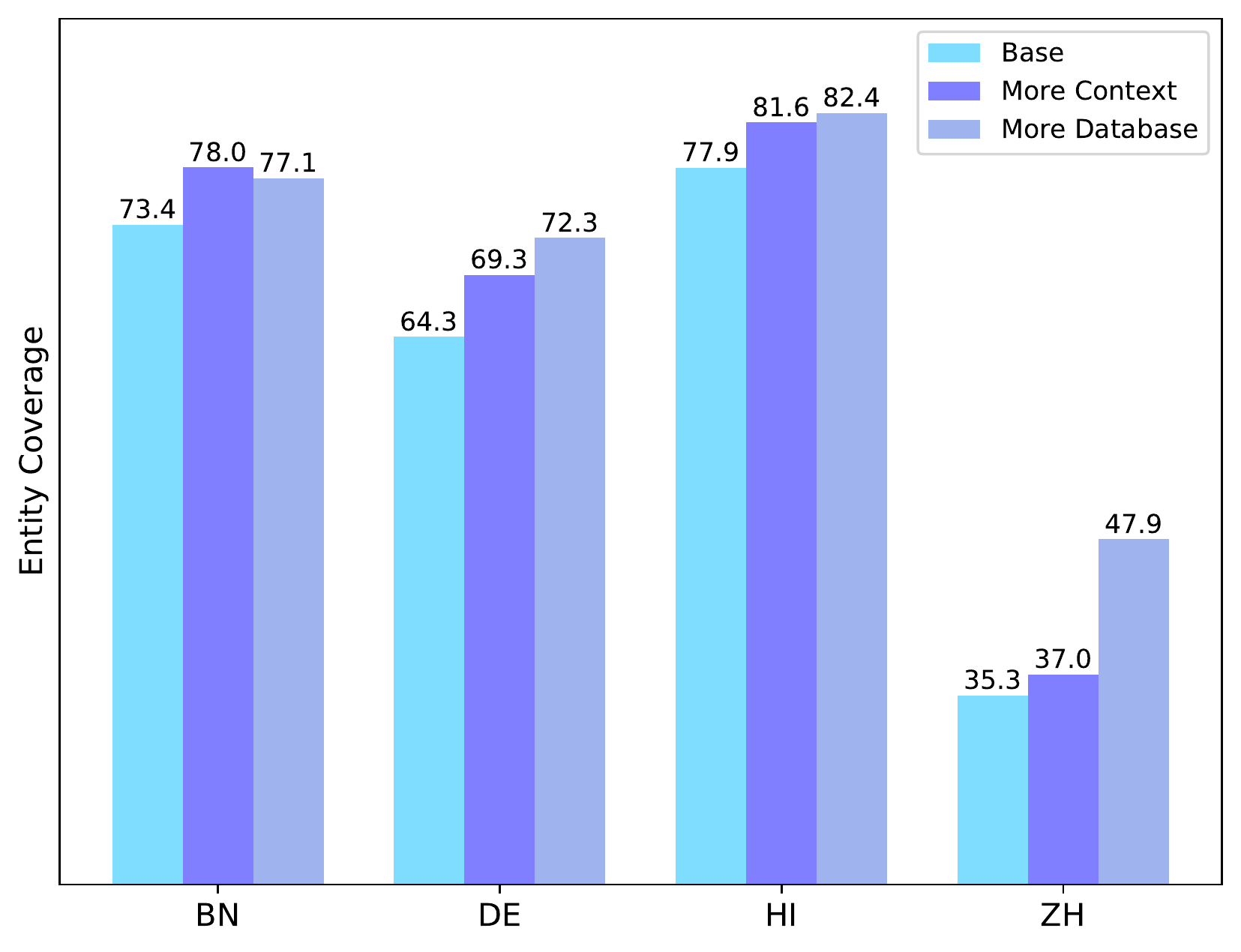}
  \caption{Entity coverage of the retrieval context for the annotated entities within the query sentence. 
%on \texttt{Base}, \texttt{More Context} and \texttt{More Database} three settings.\
}
  \label{fig:entity}
\end{figure}

\begin{table*}[h!]
\centering
\small
% \scalebox{0.9}{
\scalebox{1.0}{
\renewcommand\arraystretch{0.7}
\begin{tabular}{lcc}
    \toprule
        \textbf{Retrieval Strategy} & \textbf{Query} & \textbf{Retrieval Result}  \\
        \midrule
        \tabincell{l}{\texttt{TEXT2TEXT}}  & 
        \tabincell{c}{from 1995 to 2011 deal hudson \\ was the magazine's publisher.} & 
        \tabincell{c}{1. In 1995 Hudson became publisher of the conservative \\ Roman Catholic magazine, Crisis. \\ 2. Hudson is the former publisher and editor of \\ 3. Hudson also hosts the radio show Church and Culture \\ on Ave Maria Radio \\ ...}  \\ 
        \midrule
        \tabincell{l}{\texttt{TEXT2ENT}}  &
        \tabincell{c}{from 1995 to 2011 deal hudson \\ was the magazine's publisher.} & 
        \tabincell{c}{1. Deal W. Hudson \\ 2. Deal Wyatt Hudson \\ 3. S. Hudson \\ ...}  \\
        \midrule
        \tabincell{l}{\texttt{ENT2ENT}}  &
        \tabincell{c}{[deal hudson]} & 
        \tabincell{c}{Type: human \\ Description: Hudson is the former publisher and editor \\ of Crisis Magazine and InsideCatholic.com.}  \\
        \bottomrule
    \end{tabular}}
\caption{Examples of different retrieval strategies related to the input sentence: \textit{"from 1995 to 2011 deal hudson was the magazine's publisher."} with its corresponding entity \textit{"deal hudson"}.}
\label{tab:retrieval_augmentation_examples}
\end{table*}

While \citet{chen-etal-2022-ustc} uses Wikidata to build their gazetteer, we explore to enhance our retrieval system with Wikidata.
Wikidata is a free and entity-centric knowledge base. Every entity of Wikidata has a page consisting of a label, several aliases, descriptions, and one or more entity types.  
As shown in Figure \ref{fig:entity}, \texttt{Base} indicates that only the Wikipedia knowledge base is used, and \texttt{More Database} indicates that we use both Wikipedia and Wikidata knowledge bases. 
The entity coverage improves on all 4 languages and achieves the maximum gain of 12.6\% on \texttt{ZH}.
In addition, as \texttt{More Context} shows, expanding the length of the retrieval context also brings more entity knowledge.
Thus, we use the infusion approach to make more retrieval context visible to model. More details are described in Section \cref{4.2}.

\section{Methodology} \label{sec:methods}

\paragraph{Overview} 
\begin{figure*}
\centering
\includegraphics[width=0.9\textwidth]{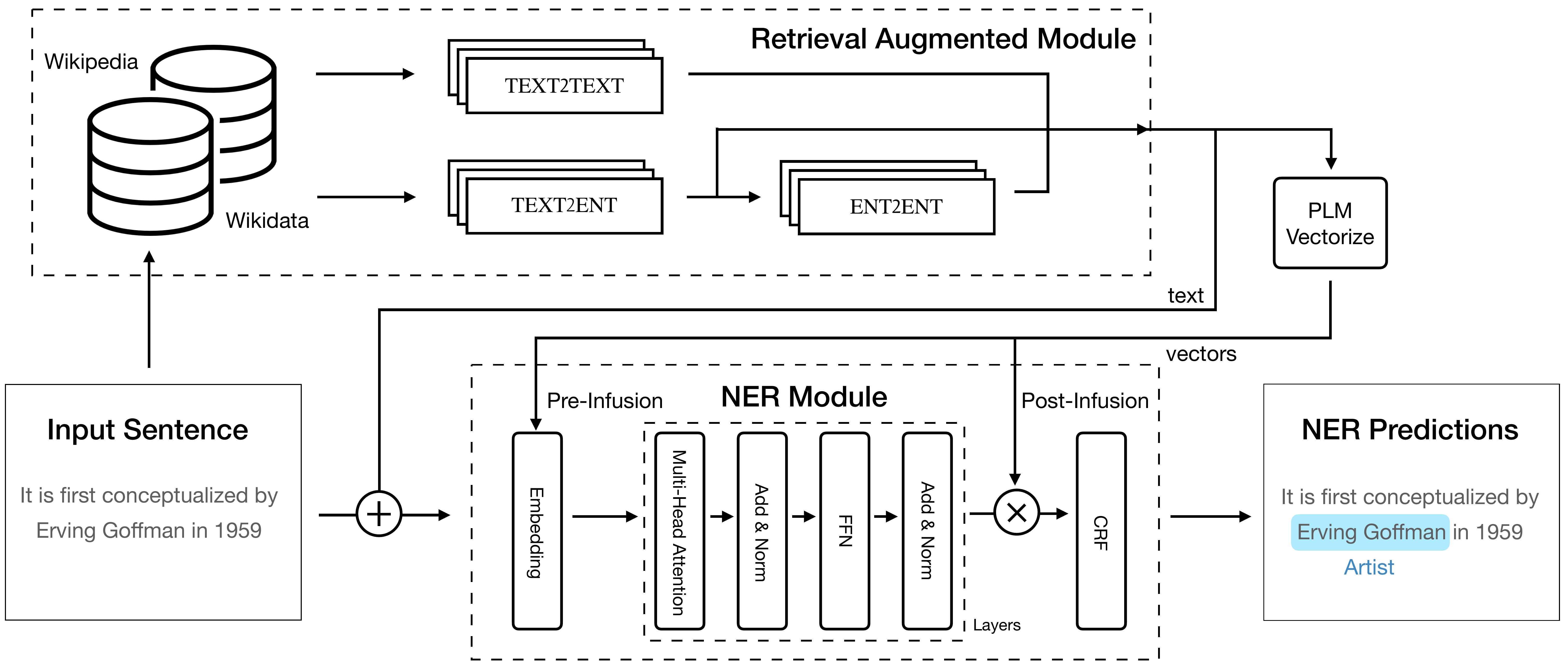}
\caption{Overall architecture of U-RaNER.} \label{fig:arch}
\end{figure*}

As depicted in Fig.~\ref{fig:arch}, U-RaNER is comprised of two parts: a retrieval augmentation module and a NER module. The retrieval augmentation module utilizes multiple retrieval strategies and the NER module adopts a modified transformer structure to utilize the retrieved knowledge.
Given an input sentence, U-RaNER retrieves similar texts and entities as external knowledge, which are then utilized in the form of text and vectors to help the NER module obtain improved predictions.

\subsection{Retrieval Augmentation Module} \label{sec:4.1}
In the retrieval augmentation module, we design three different retrieval strategies, namely \texttt{TEXT2TEXT}, \texttt{TEXT2ENT}, and \texttt{ENT2ENT}, which aim to obtain a variety of useful information from different sources to enhance our NER model.

\paragraph{\texttt{TEXT2TEXT}} 
The \texttt{TEXT2TEXT} retrieval strategy is to obtain texts related to input sentences from Wikipedia by the way of sparse retrieval~\cite{DBLP:journals/cj/McDonell77, DBLP:journals/ftir/RobertsonZ09}. 
Through this form of retrieval, the goal is to obtain additional and useful relevant information as much as possible to alleviate the low-context problem of MultiCoNER.
Specifically, we first parse the latest Wikipedia dumps and use ElasticSearch~\footnote{\url{https://github.com/elastic/elasticsearch}} to index them. 
And finally, we use each sentence in the dataset as the query and use the BM25 retrieval algorithm that comes with ElasticSearch to search in the built index database to obtain the Top-K documents related to the input sentence from Wikipedia, as shown in the first example of Table~\ref{tab:retrieval_augmentation_examples}. Note that the \texttt{TEXT2TEXT} strategy is used by \citet{wang-etal-2022-damo} to win 10 out of 13 tracks when competing in the SemEval-2022 Task 11.

\paragraph{\texttt{TEXT2ENT}} 
The \texttt{TEXT2ENT} retrieval strategy aims to retrieve candidate entities that may be mentioned in input sentences, as illustrated in the second example of Table~\ref{tab:retrieval_augmentation_examples}.
We believe that if the candidate entities that may be mentioned in the sentence can be retrieved in advance, the related knowledge might be helpful to build a stronger entity recognition model.
The \texttt{TEXT2ENT} strategy is inspired by the related technologies of dictionary disambiguation~\cite{harige-buitelaar-2016-generating} and  entity linking~\cite{DBLP:conf/iclr/CaoI0P21}.
But dictionary disambiguation can only perform hard matching, and there is no detailed annotation information for entity linking (that is, the corresponding information between span and entity), so these two traditional methods cannot be directly applied to our scene.
Therefore, in this part of the specific practice, we tried two different retrieval methods, namely sparse retrieval and dense retrieval. 
The details of these two retrieval methods are in the Appendix \ref{app:text2ent}.

\paragraph{\texttt{ENT2ENT}} 
The \texttt{ENT2ENT} retrieval strategy aims to retrieve some entities and their corresponding information from Wikidata. 
Wikidata integrates billions of structural information between millions of entities, such as the alias of entities and the relationships of entity pairs. And intuitively, such information is beneficial to our NER model. 

In the process of \texttt{ENT2ENT} retrieval, we want to find out external entity types which maybe inspire the entity labeling of the input sentence. Concretely, for each given entity, we first retrieve Wikidata to get its relevant Wikidata entities. Next, we gather and utilize the properties of the Wikidata entities from their corresponding Wikidata pages. In particular, we take the ``instance of" and ``sub-class of" properties as the entity types. For example, as shown in Table~\ref{tab:retrieval_augmentation_examples}, with entity ``deal hudson'' as the query, \texttt{ENT2ENT} strategy will retrieve its type (i.e., ``human'') and description text.
Finally, all relevant Wikidata entities and their types are as the retrieved augmented data.
The detailed procedure for \texttt{ENT2ENT} is in the Appendix \ref{app:ent2ent}.

\subsection{Named Entity Recognition Module}
\label{sec:4.2}
\paragraph{BERT-CRF}
\label{sec:bert-crf}
We use \textit{xlm-roberta-large} (XLM-R) \citep{conneau-etal-2020-unsupervised} as the PLMs for all the tracks.
Given an input sentence $\mathbf{x} = x_1, x_2, \ldots, x_{n}$,
% ($x_i$ is the i-th token, $n$ is the sentence length), 
transformer-based standard fine-tuning for NER first feeds the input sentence $\mathbf{x}$ into the PLMs to get the token representations $\mathbf{h}$.
The token representations $\mathbf{h}$ are fed into a CRF layer to get the conditional probability $p_{\theta}(\mathbf{y} \mid \mathbf{h})$, and the model is trained by maximizing the conditional probability and minimizing the cross entropy loss:
% $\mathcal{L}=-\sum_{i=1}^{n} \log p_{\theta}(\mathbf{y} \mid \mathbf{h})$.
$\mathcal{L}=-\log p_{\theta}(\mathbf{y} \mid \mathbf{h})$.

\paragraph{RaNER}
\label{sec:raner}
Given the retrieval context $\mathbf{\tilde{x}}$, we define a neural network parameterized by $\theta$ that learns from a concatenated input $[\mathbf{x} ; \tilde{\mathbf{x}}]$.
We feed the input and retrieve the representation $[\mathbf{h} ; \tilde{\mathbf{h}}]$:
\begin{equation}
\scalebox{.8}{$
[\mathbf{h} ; \tilde{\mathbf{h}}] = [h^{(1)}, \dots h^{(n)}, \tilde{h}^{(1)}, \dots \tilde{h}^{(n)}]=\operatorname{embed}([\mathbf{x} ; \tilde{\mathbf{x}}])$
}
\end{equation}
We then feed $\mathbf{h}$ into the CRF layer and train by minimizing the conditional probability $p_{\theta}(\mathbf{y} \mid \mathbf{h})$ as mentioned above.

\paragraph{U-RaNER}
\label{4.2}
To exploit more retrieval contexts, we first slice $\tilde{\mathbf{x}}$ by model-limited input length as $\tilde{\mathbf{x}} = \tilde{x}_0, \tilde{x}_1, \ldots, \tilde{x}_{m}$.
Then, we keep $\tilde{x}_0$ as the text for concatenation, and feed the rest context list into PLM as $[(\mathbf{x} ; \tilde{x}_{1}), \ldots, (\mathbf{x} ; \tilde{x}_{m})]$, 
which is used in \citet{lewis2020retrieval} for better information interaction, 
and get the token vector list $[(\mathbf{h}_{1} ; \tilde{h}_{1}), \ldots, (\mathbf{h}_{m} ; \tilde{h}_{m})]$. 
Afterwards, we consider two infusion (Pre-Infusion and Post-Infusion) approaches using the representation $[\tilde{h}_{1}, \ldots, \tilde{h}_{m}]$ and $[\mathbf{h}_{1}, \ldots, \mathbf{h}_{m}]$, respectively.

For \texttt{Pre-Infusion}, we fetch the token vectors of the corresponding positions of the anchors from the vector list $[\tilde{h}_{1}, \ldots, \tilde{h}_{m}]$.
Then, we perform the mean operation to obtain the set of anchor vectors $\mathcal{V} \in \mathbb{R}^{p \times d}$, $p$ is the number of anchors, and $d$ is the hidden size.
% $\mathcal{V} = v_1, v_2, \ldots, v_{p}$ ($\mathcal{V} \in \mathbb{R}^{p \times d}$, $p$ is the number of entities, and $d$ is the embedding size),
Considering that the word embedding layer in XLM-R has two input modes, including vocabulary index input as well as word embedding input, we first perform the former for $[\mathbf{x};\tilde{x}_{0}]$ to obtain the input text embedding $E$, and later concatenate $E$ and the anchor vectors $\mathcal{V}$ to form the word embedding input. 
Finally, we get the representation $[\mathbf{h} ; \tilde{h}_{0}; \tilde{h}_{v}]$. We only use $\mathbf{h}$ to pass the CRF layer.

For \texttt{Post-Infusion}, we first feed $[\mathbf{x};\tilde{x}_{0}]$ to XLM-R and get the token representation $[\mathbf{h} ; \tilde{h}_{0}]$.
For input representation list $[\mathbf{h} ; \mathbf{h}_{1}, \ldots, \mathbf{h}_{m}]$, 
we perform the max operation on the token dimension to obtain the final representation $\mathbf{h}_{\max}$. Then, we use $\mathbf{h}_{\max}$ for calculation as in \texttt{BERT-CRF}.
Notably, we find that the post-infusion method is superior to the pre-infusion method in our preliminary experiments, and the default infusion method in the experimental section is post-infusion.

\subsection{Ensemble Module}
Given predictions $\{\hat{\vy}_{\theta_1}, \cdots, \hat{\vy}_{\theta_m}\}$ from $m$ models with different random seeds, we use majority voting to generate the final prediction $\hat{\vy}$. 
% We convert the label sequences into entity spans to perform majority voting. 
Following \citet{yamada-etal-2020-luke,wang-etal-2022-damo}, the module ranks all spans in the predictions by the number of votes in descending order and selects the spans with more than 50\% votes into the final prediction. The spans with more votes are kept if the selected spans have overlaps and the longer spans are kept if the spans have the same votes.

\begin{table*}[t!]
\centering
\setlength{\tabcolsep}{2mm}
\scalebox{0.75}{
\begin{tabular}{lccccccccccccc|c}
\toprule
System 	& \textbf{EN} & \textbf{ES} & \textbf{SV} & \textbf{UK} & \textbf{PT} & \textbf{FR} & \textbf{FA} & \textbf{DE} & \textbf{ZH} & \textbf{HI} & \textbf{BN} & \textbf{IT} & \textbf{MULTI} & \textbf{AVG.} \\
\midrule
        BERT-CRF & 62.80  & 65.34  & 68.68  & 67.68  & 64.37  & 66.05  & 60.70  & 69.44  & 62.02  & 73.08  & 71.82  & 68.15  & 63.27  & 66.42 \\
        NLPeople & 71.81  & 72.76  & 75.08  & 73.41  & 70.16  & 72.85  & 70.76  & 77.67  & 65.96  & 78.50  & 78.24  & 73.71  & 78.38  & 73.79 \\ 
        USTC-NELSLIP & 72.15  & 74.44  & 75.47  & 74.37  & 71.26  & 74.25  & 68.85  & 78.71  & 66.57  & 82.14  & 80.59  & 75.70  & 75.62  & 74.62  \\ 
        IXA/Cogcomp & 72.82  & 73.81  & 76.54  & 75.25  & 72.28  & 74.52  & 69.49  & 80.35  & 64.86  & 79.56  & 78.95  & 74.67  & 78.17  & 74.71  \\ 
        CAIR-NLP & 79.33  & 83.63  & 82.88  & 81.29  & 80.16  & 83.08  & 77.50  & 74.71  & 58.43  & 72.23  & 69.46  & 83.78  & 79.16  & 77.36  \\ 
        PAI & 80.00  & 71.67  & 72.38  & 71.28  & 81.61  & 86.17  & 68.46  & \textbf{88.09}  & 74.87  & \textbf{80.96}  & \textbf{84.39}  & 84.88  & 77.00  & 78.60  \\
        % Samsung Research & 83.09 & - & - & - & - & - & - & - & -  & - & - & - & - & - \\
        NetEase.AI & - & - & - & - & - & - & - & - & \textbf{84.05}  & - & - & - & - & - \\ 
    \rowcolor[HTML]{EEF6FF}    Ours & \textbf{85.53}  & \textbf{89.78}  & \textbf{89.57}  & \textbf{89.02}  & \textbf{85.97}  & \textbf{89.59}  & \textbf{87.93}  & 84.97  & 75.98  & 78.56  & 81.60  & \textbf{89.79}  & \textbf{84.48}  & \textbf{85.60} \\
\bottomrule
\end{tabular}
}
\caption{Part of the official results on the leaderboard. \texttt{BERT-CRF} is the post-evaluation results of our baseline system (BERT-CRF) on the released test set.}
\label{tab:results_test}
\end{table*}

\section{Experimental Setup} \label{sec:exp}
\subsection{Datasets and Evaluation Metrics}
We use the official MultiCoNER \RNum{2} dataset \citep{multiconer2-data} in all tracks to train our models. 
The detailed data statistics is in the Appendix \ref{app:multi} and \ref{app:taxonomy}.
% The multilingual NER corpus (MultiCoNER \footnote{\url{https://multiconer.github.io/dataset}}) aims to recognize the complex named entities, like the titles of creative works which are not simple nouns, and pose challenges for current NER systems. 
% With the same set of tags, the 12 multilingual datasets specifically include:
% BN-Bangla, DE-German, EN-English, ES-Spanish, FA-Farsi,  FR-French,
% HI-Hindi, IT-Italian, PT-Portuguese, SV-Swedish, UK-Ukrainian and ZH-Chinese.
% Table \ref{tab:multiconer} shows the detailed dataset statistics.
The results on the leaderboard are evaluated with the entity-level macro F1 scores, which treat all the labels equally \footnote{In comparison, most of the publicly available NER datasets (e.g., CoNLL 2002, 2003 datasets) are evaluated with the entity-level micro F1 scores, which emphasize common labels \citep{akbik-etal-2018-contextual,devlin-etal-2019-bert,yamada-etal-2020-luke,wang-etal-2022-damo}. 
Except for the results in Table \ref{tab:results_test}, the following results are entity-level micro F1 scores if not otherwise specified.}. 

\subsection{Training Strategy}

\paragraph{NER Model Training}
% For fair comparison with prior systems, we use \textit{xlm-roberta-large} \citep{conneau-etal-2020-unsupervised} as our initial checkpoint. 
% We use the AdamW \citep{DBLP:journals/corr/abs-1711-05101} optimizer with a linear warmup-decay learning schedule and a dropout \citep{srivastava2014dropout} of 0.1. 
% We set the batch size and learning rate to 16 and 2e-5, and train models over 4 random seeds.
% According to the dataset sizes, we train the models for 5 epochs and 20 epochs for multilingual and monolingual models respectively. 
Our final NER models are trained on the combined dataset including both the training and development sets on each track to fully utilize the labeled data. 
% For models trained on the training set, we use the best macro F1 on the development set during training to select the best model checkpoint. 
For models trained on the combined dataset, we use the final model checkpoint after training.
The detailed system configurations is in the Appendix \ref{app:system}

\paragraph{Multi-stage Fine-tuning}
Multi-stage fine-tuning (MSF) aims at transferring the parameters of fine-tuned embeddings in a model at an early stage into other models in the next stage \citet{shi-lee-2021-tgif}. 
The approach stores the checkpoint of fine-tuned XLM-R embeddings at the early stage and uses it as the initialization of XLM-R embeddings for model training at the next stage. 
\citet{wang-etal-2022-damo} experimentally demonstrates that MSF can leverage the annotations from all tracks and thus improve performance and accelerate training.
In addition, we observe that inconsistent training set sizes on different language tracks can also lead to degradation of model performance. We use increasing batch size and upsampling strategy to address this issue. The details are shown in the Appendix \ref{app:msf}.

\subsection{Baselines}
In this paper, we compare the proposed U-RaNER with the following baseline models:

\begin{itemize}
    \item \textbf{BERT-CRF}, as introduced in \ref{sec:bert-crf}, is composed of a BERT-like encoder and a CRF decoder . It is widely used for sequence labeling tasks.  We use \textit{xlm-roberta-large} (XLM-R) \citep{conneau-etal-2020-unsupervised} as the pretrained backbone for all the tracks.
    \item \textbf{RaNER}, as introduced in \ref{sec:raner}, improves BERT-CRF by incorporating retrieval contexts as input for better performance. Retrieval augmented methods have proven to be highly effective in the NER task\cite{wang-etal-2021-improving,zhang-etal-2022-domain,wang-etal-2022-named}.
    \item \textbf{RaNER-MSF} \cite{wang-etal-2022-damo} achieves the previous best overall performance on the Multi-CoNER \RNum{1} dataset, which exploits multi-stage fine-tuning to leverage the annotations from all tracks and thus improve performance and accelerate training of RaNER.
    % \item \textbf{ChatGPT}\footnote{\url{https://openai.com/blog/chatgpt/}}, 
    \item \textbf{ChatGPT}, also known as \textit{gpt-3.5-turbo}, is the most capable GPT-3.5 \cite{ouyang2022training} model and optimized for chat.
    Following \citep{lai2023chatgpt}, our prompt structure for ChatGPT consists of a task description, a note for output format, and an input sentence. Despite a \texttt{Single-turn} prompting strategy, we additionally try two enhanced prompting strategies: \texttt{Multi-turn} and \texttt{Multi-ICL}. \texttt{Multi-turn} first performs the task in 6 coarse-grained categories, and later performs finer-grained NER. \texttt{Multi-ICL} constructs demonstrations spliced after the note part by randomly selecting examples from the training set.
    The detailed prompting procedure for \texttt{Single-turn}, \texttt{Multi-turn} and \texttt{Multi-ICL} is in the Appendix \ref{app:prompt}.
    
\end{itemize}

\section{Results and Analysis}
\label{sec:results}
% In this section, we use language codes\footnote{\url{https://en.wikipedia.org/wiki/List_of_ISO_639-1_codes}} to represent languages, and use \textsc{\textbf{multi}} to represent multilingual track.

\begin{table*}[ht!]
\centering
\scalebox{0.68}{
\begin{tabular}{lccccccccccccccc|c}
\toprule
        Method & $\triangle$ & $\dagger$ & $\ddagger$ & \textbf{BN} & \textbf{DE} & \textbf{EN} & \textbf{ES} & \textbf{FA} & \textbf{FR} & \textbf{HI} & \textbf{IT} & \textbf{PT} & \textbf{SV} & \textbf{UK} & \textbf{ZH} & \textbf{AVG.} \\
        % Method & Infusion & Wikipedia & Wikidata & \textbf{BN} & \textbf{DE} & \textbf{EN} & \textbf{ES} & \textbf{FA} & \textbf{FR} & \textbf{HI} & \textbf{IT} & \textbf{PT} & \textbf{SV} & \textbf{UK} & \textbf{ZH} & \textbf{AVG.} \\
\midrule
ChatGPT w/ \\
        \hspace{1mm}\texttt{Single-turn} & \xmark & \xmark & \xmark & 7.24  & 10.06  & 13.36  & 12.44  & 10.94  & 11.05  & 9.04  & 16.32  & 17.27  & 18.03  & 10.88  & 5.02  & 11.80  \\ 
        \hspace{1mm}\texttt{Multi-turn} & \xmark & \xmark & \xmark & 8.12  & 14.57  & 15.38  & 15.52  & 12.75  & 13.60  & 9.17  & 17.81  & 17.70  & 20.38  & 14.25  & 5.60  & 13.74  \\ 
        \hspace{1mm}\texttt{Multi-ICL} & \xmark & \xmark & \xmark & 9.76  & 14.84  & 17.65  & 16.28  & 14.11  & 13.95  & 10.48  & 18.63  & 18.84  & 20.94  & 15.57  & 6.34  & 14.78  \\ 
\midrule
        BERT-CRF & \xmark & \xmark & \xmark & 86.98  & 76.08  & 72.61  & 75.66  & 69.37  & 74.44  & 85.46  & 80.70  & 76.54  & 78.48  & 76.30  & 75.11  & 77.31  \\ 
        RaNER & \xmark & \cmark & \xmark & 92.30  & 84.29  & 84.32  & 88.81  & 87.85  & 86.77  & 91.75  & 91.08  & 88.45  & 89.74  & 88.46  & 81.55  & 87.95  \\ 
        RaNER-MSF & \xmark & \cmark & \xmark & 93.11  & 86.81  & 86.82  & 90.90  & 89.52  & 88.99  & 93.97  & 92.42  & 90.75  & 91.93  & 90.93  & 82.83  & 89.92  \\ 
\midrule
U-RaNER w/ \\
        \hspace{1mm}\texttt{TEXT2ENT$^\star$} & \xmark & \xmark & \cmark & 89.87  & 85.83  & 87.54  & 88.03  & 86.44  & 83.86  & 86.82  & 91.19  & 78.92  & 86.20  & 84.26  & 85.62  & 86.22  \\ 
        \hspace{1mm}\texttt{ENT2ENT$^\star$} & \xmark & \xmark & \cmark & 94.45  & 88.85  & 88.11  & 91.34  & 89.70  & 89.96  & 94.68  & 91.53  & 90.15  & 91.68  & 88.21  & 87.02  & 90.47 \\
        \hspace{1mm}\texttt{TEXT2TEXT} & \cmark & \cmark & \xmark & 94.36  & 87.79  & 88.07  & 92.57  & 90.91  & 91.80  & 94.25  & 93.60  & 91.94  & 93.02  & 91.40  & 84.11  & 91.15  \\ 
        \hspace{1mm}\texttt{TEXT2ENT} & \cmark & \cmark & \cmark & 94.77  & 89.48  & 89.88  & 93.46  & 90.80  & 90.83  & 94.57  & 93.83  & 92.12  & 93.20  & 91.12  & 89.41  & 91.96  \\ 
        \hspace{1mm}\texttt{ENT2ENT} & \cmark & \cmark & \cmark & \textbf{94.96 } & \textbf{90.36 } & \textbf{90.62 } & \textbf{93.51 } & \textbf{91.85 } & \textbf{92.88 } & \textbf{95.12 } & \textbf{94.60 } & \textbf{92.90 } & \textbf{94.45 } & \textbf{91.57 } & \textbf{90.38 } & \textbf{92.77} \\
\bottomrule
\end{tabular}
}
\caption{ 
The top bar shows ChatGPT's performance (micro-F1 scores) using three prompting strategies, the former two being zero-shot learning and \texttt{Multi-ICL} being few-shot learning.
Following the comparison between the top system \citep{wang-etal-2022-damo} in the MultiCoNER \RNum{1} and the three variants of our method on the validation set.
$\star$ indicates that we merely use the Wikidata knowledge base.
$\triangle$ means we scale the model horizon with the infusion approach.
$\dagger$ and $\ddagger$ indicate the use of the Wikipedia or Wikidata knowledge base.}
\label{tab:results_dev}
\end{table*}

\subsection{Main Results}
There are 45 teams that participated in the MultiCoNER \RNum{2} shared task. Due to limited space, we only compare our system with the systems from teams NLPeople, USTC-NELSLIP, IXA/Cogcomp, CAIR-NLP, PAI and NetEase.AI\footnote{Please refer to \url{https://multiconer.github.io/results} for more details about the results.}. As NetEase.AI solely took part in the Chinese track, which means we only have access to their results for this specific track.
In the post-evaluation phase, we evaluate the baseline system without the use of additional knowledge bases to further show the effectiveness of our retrieval-augmented system. 
The official results and the results of our baseline system are shown in Table \ref{tab:results_test}. 
Our system performs the best on 9 out of 13 tracks with the average result exceeding the second-place system by the absolute F1-measure of 7.0\%.
Moreover, our system outperforms our baseline by the 19.18\% F1-measure on average, 
which demonstrates that the retrieval-augmented system based on multiple knowledge bases is extremely helpful in identifying complex entities, leading to significant improvement on model performance.

In addition, we use three prompting strategies to evaluate ChatGPT. Due to the overwhelming number of test sets (millions of levels), the expense of invoking the OpenAI interface is unaffordable. We experiment on the validation set and the results are in Table \ref{tab:results_dev}.
We observe that ChatGPT's performance on the multilingual NER dataset is quite poor, with an average F1-score of only 14.78\% by the best strategy.
Even on the coarse-grained level the result is merely 29.70\% (Table \ref{tab:type_coarse}), which is comparable to the result measured on MultiCoNER \RNum{1} \citep{multiconer-report} by \citet{lai2023chatgpt}.

\subsection{Ablation Study}

In this section, we perform extensive ablation experiments to show the effectiveness of various settings in our retrieval-augmented system.
Following \citet{wang-etal-2022-damo}, we employ the multi-stage fine-tuning (MSF) training strategy.
As shown in Table \ref{tab:results_dev}, the model performance improves from 87.95\% to 89.92\%, which illustrates the effectiveness of the multi-stage training.
Note that the following five rows in Table \ref{tab:results_dev} all use the MSF training strategy.

For the different knowledge sources, the use of Wikipedia data achieves the gain of 12.61\% (\texttt{RaNER-MSF} vs. \texttt{BERT-CRF}), the use of wikidata data achieves the gain of 13.16\% (\texttt{ENT2ENT$^\star$} vs. \texttt{BERT-CRF}), and using both together achieves the maximum gain of 15.46\% (\texttt{ENT2ENT} vs. \texttt{BERT-CRF}). 
This shows that knowledge is highly useful for system performance and illustrates the complementarity of the two knowledge bases.

For the different knowledge acquisition methods, the \texttt{ENT2ENT} approach is superior to the \texttt{TEXT2ENT} approach (90.47\% vs. 86.22\%).
In addition, we use the infusion approach to further improve the model performance (\texttt{RaNER-MSF} vs. \texttt{TEXT2TEXT}), which suggests that guaranteeing knowledge to be visible to model is also important.
The default infusion method in our experiments is post-infusion. We also analyze the impact of the two different infusion methods on performance in the Appendix \ref{app:infuse}.

\subsection{Coarse-and-fine Category Analysis}

\begin{table}[t!]
\centering
\setlength{\tabcolsep}{1.8mm}
\scalebox{0.73}
{
\renewcommand\arraystretch{1.1}
\begin{tabular}{clccccc|c}
\toprule
         ~~~~~ & \textbf{Method} & \textbf{BN} & \textbf{ES} & \textbf{PT} & \textbf{SV} & \textbf{ZH} & \textbf{AVG.} \\ 
\midrule
        \parbox[t]{2mm}{\multirow{4}{*}{\rotatebox[origin=c]{90}{Coarse}}} &
        ChatGPT & 21.26  & 33.86  & 35.27  & 40.11  & 18.01  & 29.70 \\
        \cmidrule{2-8}
        & RaNER & 95.92  & 96.17  & 96.79  & 97.55  & 91.94  & 95.67  \\ 
        & U-RaNER & 97.48  & 98.30  & 98.33  & 98.49  & 95.55  & \textbf{97.63}  \\ 
        % & \rowcolor[HTML]{EEF5FF} $\Delta$ & +1.56  & +2.13  & +1.54  & +0.94  & +3.61  & +1.96  \\ 
        & $\Delta$ & +1.56  & +2.13  & +1.54  & +0.94  & +3.61  & +1.96  \\ 
\midrule
        \parbox[t]{2mm}{\multirow{4}{*}{\rotatebox[origin=c]{90}{Fine}}} &
        ChatGPT & 9.76  & 16.28  & 18.84  & 20.94  & 6.34  & 14.43  \\
        \cmidrule{2-8}
        & RaNER & 93.11  & 90.90  & 90.75  & 91.93  & 82.83  & 89.90  \\ 
        & U-RaNER & 94.96  & 93.51  & 92.90  & 94.45  & 90.38  & \textbf{93.24}  \\ 
        % & \rowcolor[HTML]{EEF5FF} $\Delta$ & +1.85  & +2.61  & +2.15  & +2.52  & +7.55  & +3.34 \\ 
        & $\Delta$ & +1.85  & +2.61  & +2.15  & +2.52  & +7.55  & +3.34 \\ 
\bottomrule
\end{tabular}
}
\caption{The performance comparison between RaNER and U-RaNER at coarse-and-fine grained categories.}
\label{tab:type_coarse}
\end{table}

To illustrate the advantages of U-RaNER on fine-grained NER, we transform the model predictions to the coarse-grained level according to the official topology of fine-grained categories.
We use the models of RaNER-MSF and U-RaNER w/ \texttt{ENT2ENT} in Table \ref{tab:results_dev}  for the analysis.
As shown in the Table \ref{tab:type_coarse}, the improvements in coarse-grained metrics are significantly lower than those of fine-grained metrics, differing by 1.38\% (3.91\% on the \texttt{ZH} track).
It suggests that the proposed U-RaNER is better at coping with complex scenarios of fine-grained classification.
Besides, the average F1 for ChatGPT at different granularity is significant distinct (29.70\% vs. 14.43\%), which shows the difficulty in identifying fine-grained complex entities.

\subsection{Query Relevance}

\begin{figure}[t]
    \centering
    \includegraphics[width=0.85\linewidth]{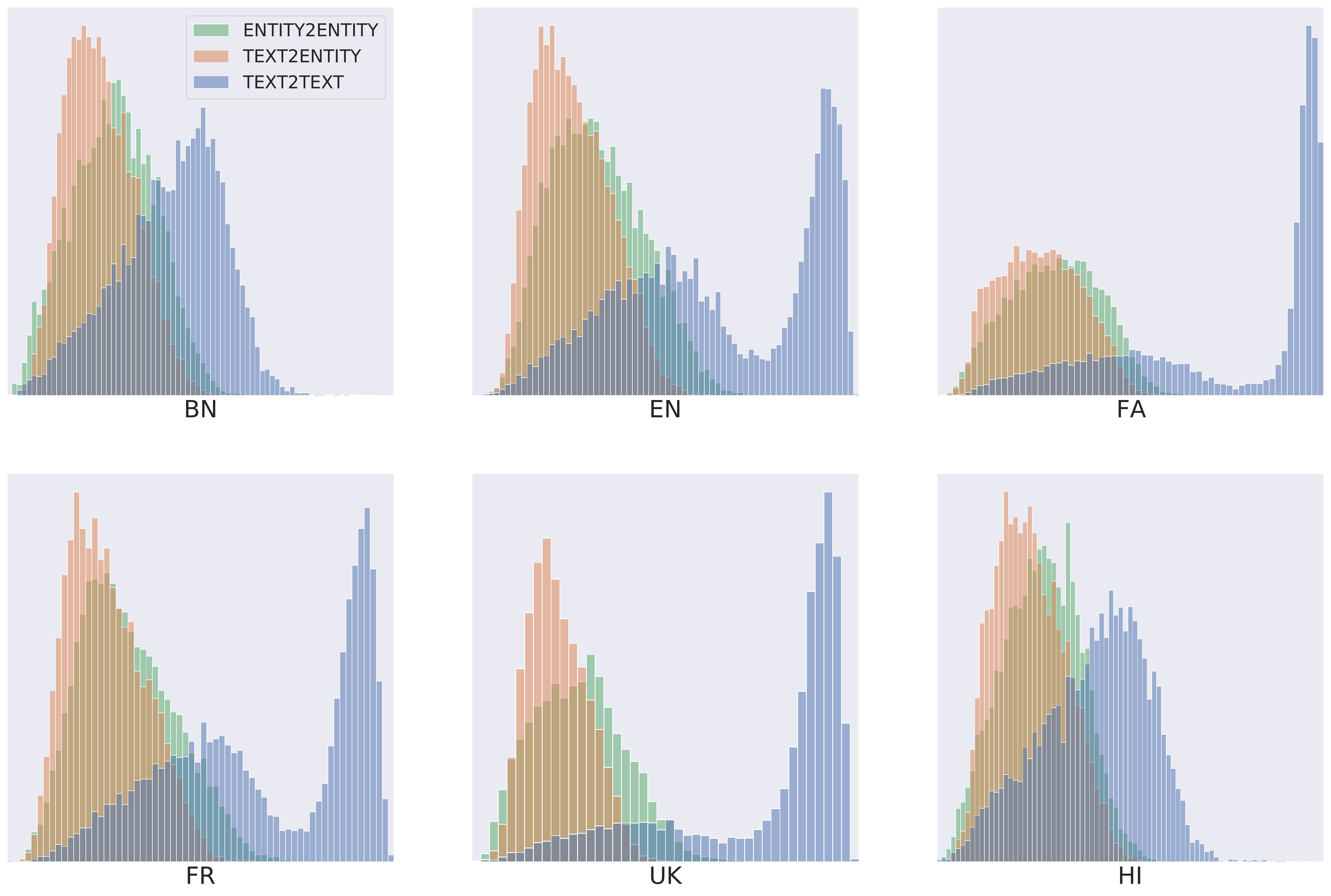}
    \caption{The distribution of the character-level IoU between query and its retrieval result. Each subplot is the histograms of different retrieval strategies on the corresponding dataset, where the $x$-axis indicates the IoU values ranging from 0 to 1. }
    \label{fig:iou}
\end{figure}

We define a relevance metric to compute the relevance between the query and retrieval result. The metric calculates the Intersection-over-Union (IoU) between the characters \footnote{We take repeat characters as different characters.} of the query and those of the retrieved result. We plot the results on the training set of 6 tracks in Figure~\ref{fig:iou}. 
It can be observed that the IoU values of \texttt{TEXT2TEXT} strategy form a larger cluster than those of \texttt{TEXT2ENT} and \texttt{ENT2ENT}, which indicates that \texttt{TEXT2TEXT} retrieval would focus more on the context instead of merely the entities in the query text. Additionally, we observe that the distributions of \texttt{ENT2ENT} have larger medians than those of \texttt{TEXT2ENT}. This might due to  \texttt{ENT2ENT} would retrieve more relevant entities from the Wikidata than \texttt{TEXT2ENT}. By employing diverse retrieval techniques, we can leverage data with distinct attributes to improve the effectiveness of the model.

\subsection{Context Length Analysis}

\begin{figure}[t]
    \centering
    \includegraphics[width=0.85\linewidth]{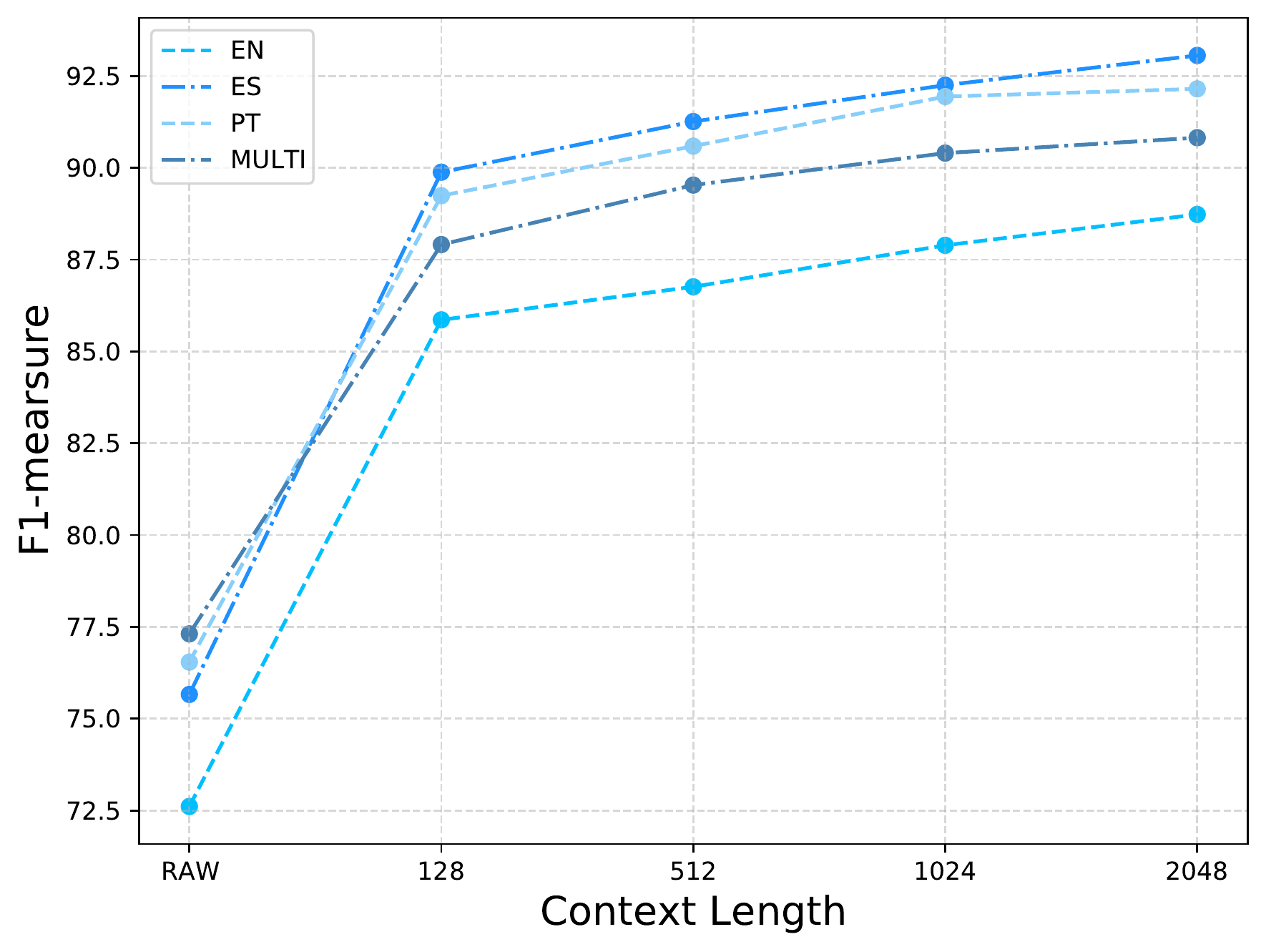}
    \caption{F1-measure with different length of context. \texttt{RAW} indicates that no external context is appended.}
    \label{fig:mi}
\end{figure}

In this section, we focus on analyzing the impact of different context length on model performance. 
We conduct a series of experiments on EN, ES, PT and MULTI datasets with the context length ranging from 128 to 2048.
We can observe from Figure \ref{fig:mi} that the model performance increases as the context length grows. However, when the context list length exceeds 1024, the trend of performance improvement on all four datasets slows down. 
This indicates that the knowledge capacity in the contexts saturates as the length of the context increases. 
For better performance, we need to find complementary and highly relevant contextual pieces as additional knowledge sources.

\subsection{Error Analysis}

\begin{table}[t!]
\centering
\small
\begin{tabular}{lccc}
    \toprule
        \textbf{Language} & \textbf{F1-entity} & \textbf{F1-mention} & \textbf{Acc-typing} \\
        \midrule
        %\textbf{BN} & \textbf{DE} & \textbf{EN} & \textbf{ES} & \textbf{FA} & \textbf{FR} & \textbf{HI} & \textbf{IT} & \textbf{PT} & \textbf{SV} & \textbf{UK} & \textbf{ZH} & \textbf{AVG.} \\
        \textbf{BN} & 92.30 & 97.33 & 94.83 \\ 
        \textbf{DE} & 84.29 & 95.00 & 88.73 \\ 
        \textbf{EN} & 84.32 & 98.15 & 85.91 \\ 
        \textbf{ES} & 88.81 & 98.13 & 90.50 \\ 
        \textbf{FA} & 87.85 & 97.21 & 90.37 \\ 
        \textbf{FR} & 86.77 & 97.34 & 89.14 \\ 
        \textbf{HI} & 91.75 & 97.15 & 94.44 \\ 
        \textbf{IT} & 91.08 & 98.53 & 92.44 \\ 
        \textbf{PT} & 88.45 & 98.45 & 89.84 \\ 
        \textbf{SV} & 89.74 & 98.60 & 91.01 \\ 
        \textbf{UK} & 88.46 & 98.33 & 89.96 \\ 
        \textbf{ZH} & 81.55 & 92.25 & 87.00 \\ \midrule
        \textbf{AVG.} & 87.84 & \textbf{97.21} & \textbf{90.35} \\ 
        \bottomrule
    \end{tabular}
\caption{Model performance of mention-detection and entity-typing on the 12 multilingual datasets.}
\label{tab:loctyp}
\end{table}

We divided the NER task into two stages: mention detection to locate entity spans, and entity typing to classify the spans with pre-defined labels. To further analyze the limitations of our proposed model, we present the experimental results on 12 languages in Table ~\ref{tab:loctyp}. The experimental results reveal that the average F1 score for mention detection is \textbf{97.21}, whereas the accuracy for entity typing is \textbf{90.35}. These results provide evidence that the bottleneck in fine-grained NER is 
% entity typing. 
typing.
% Therefore, future optimization efforts should integrate additional context-specific knowledge to enhance the typing accuracy even further.
More detailed discussion, including the different retrieval methods and case study, is in the Appendix \ref{app:retrieval} and \ref{app:case}.

\section{Conclusion}
In this paper, we propose a unified retrieval-augmented system (U-RaNER) for the MultiCoNER \RNum{2} shared task, which wins 9 out of 13 tracks in the shared task.
We expose that the bottleneck of the previous top system is the lack of knowledge.
Accordingly, we use both Wikipedia and Wikidata knowledge bases with three retrieval approaches so that more diverse knowledge can be considered.
Also, we explore the infusion approach to make more context visible to the model so as to make the best use of the resources. And the error analysis indicates that the entity typing sub-task is the bottleneck in the current system. 
% Therefore, 
% we plan to focus on optimizing the typing by designing a two-stage NER approach in our future work.
% we plan to assemble large language models such as ChatGPT to optimize fine-grained classification.
In the future, we plan to exploit the knowledge in the large language model such as ChatGPT or LLaMA by self-verification or fine-tuning some adapters, in order to achieve robust generalization performance.

\section*{Acknowledgements}
% Add any acknowledgements by uncommenting this section.
This work is supported by the Key Research and Development Program of Zhejiang Province, China (No. 2023C01152),  the Fundamental Research Funds for the Central Universities (No. 226-2023-00060), and MOE Engineering Research Center of Digital Library.

\bibliography{anthology,custom}
\bibliographystyle{acl_natbib}

% \clearpage
%% uncomment the below section to add an appendix.
\appendix
% \section{Example Appendix}
% \label{sec:appendix}

% This is a section in the appendix.

\section{Detailed Experimental Setup}
\subsection{MultiCoNER \RNum{2} Corpus}
\label{app:multi}
The multilingual NER \RNum{2} corpus (MultiCoNER \RNum{2}\footnote{\url{https://multiconer.github.io/dataset}}) aims to recognize the complex named entities, like the titles of creative works which are not simple nouns, and pose challenges for current NER systems. 
With the same set of tags, the 12 multilingual datasets specifically include:
BN-Bangla, DE-German, EN-English, ES-Spanish, FA-Farsi,  FR-French,
HI-Hindi, IT-Italian, PT-Portuguese, SV-Swedish, UK-Ukrainian and ZH-Chinese.
Table \ref{tab:multiconer} shows the detailed dataset statistics.
% Some examples from the corpus with annotations are in Figure \ref{fig:example}.

\begin{table}[ht!]
    \centering
    \small
    \begin{tabular}{lccc}
    \toprule
        Language & Training & Validataion & Test \\ 
    \midrule
        BN-Bangla & 9,708 & 507 & 19,859 \\ 
        DE-German & 9,785 & 512 & 20,145 \\ 
        EN-English & 16,778 & 871 & 249,980 \\ 
        ES-Spanish & 16,453 & 854 & 246,900 \\ 
        FA-Farsi & 16,321 & 855 & 219,168 \\ 
        FR-French & 16,548 & 857 & 249,786 \\ 
        HI-Hindi & 9,632 & 514 & 18,399 \\ 
        IT-Italian & 16,579 & 858 & 247,881 \\ 
        PT-Portuguese & 16,469 & 854 & 229,490 \\ 
        SV-Swedish & 16,363 & 856 & 231,190 \\ 
        UK-Ukrainian & 16,429 & 851 & 238,296 \\ 
        ZH-Chinese & 9,759 & 506 & 20,265 \\ 
        MUL-Multilingual & 170,824 & 8,895 & 358,668 \\ 
    \bottomrule
    \end{tabular}
\caption{Dataset statistics on MultiCoNER \RNum{2}.}
\label{tab:multiconer}
\end{table}

\subsection{System Setup}
\label{app:system}
For fair comparison with prior systems, we use \textit{xlm-roberta-large} \citep{conneau-etal-2020-unsupervised} as our initial checkpoint. 
We use the AdamW \citep{DBLP:journals/corr/abs-1711-05101} optimizer with a linear warmup-decay learning schedule and a dropout \citep{srivastava2014dropout} of 0.1. 
We set the batch size and learning rate to 16 and 2e-5, and train models over 4 random seeds.
According to the dataset sizes, we train the models for 5 epochs and 20 epochs for multilingual and monolingual models respectively. 
And all our experiments are conducted on a single NVIDIA A100 80GB GPU.
For the ensemble module, we train about 4 models for each track. 

\subsection{Fine-grained Taxonomy}
\label{app:taxonomy}

\begin{figure}[t!]
  \centering
  \includegraphics[width=0.85\linewidth]{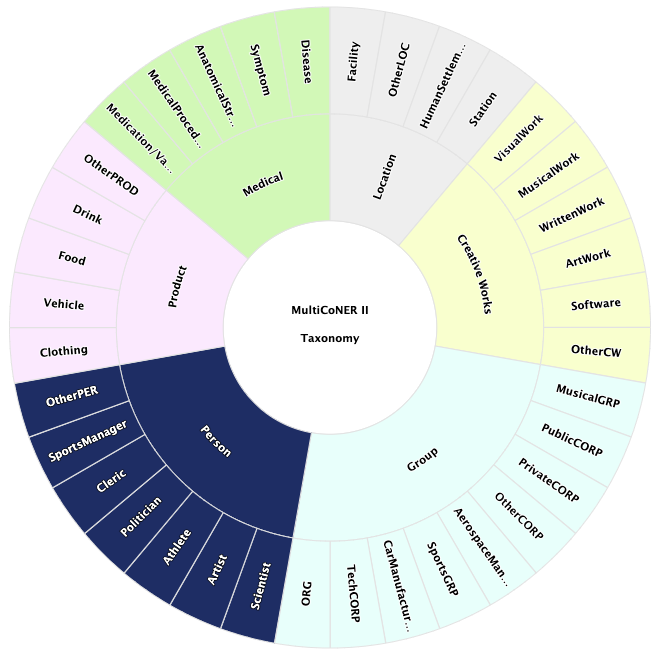}
  \caption{The taxonomy of fine-grained categories on MultiCoNER \RNum{2} from the official webpage.}
  \label{fig:taxonomy}
\end{figure}

The tagset of MultiCoNER \RNum{2} is a fine-grained tagset including 6 coarse-grained categories and 33 fine-grained categories. The coarse-to-fine mapping of the tags are as follows:
\begin{itemize}
\item Location (LOC): Facility, OtherLOC, HumanSettlement, Station;
\item Creative Work (CW): VisualWork, MusicalWork, WrittenWork, ArtWork, Software;
\item Group (GRP): MusicalGRP, PublicCORP, PrivateCORP, AerospaceManufacturer, SportsGRP, CarManufacturer, ORG;
\item Person (PER): Scientist, Artist, Athlete, Politician, Cleric, SportsManager, OtherPER;
\item Product (PROD): Clothing, Vehicle, Food, Drink, OtherPROD;
\item Medical (MED): Medication/Vaccine, MedicalProcedure, AnatomicalStructure, Symptom, Disease.
\end{itemize}
The Figure \ref{fig:taxonomy} shows the fine-grained taxonomy.
% of the corpus.

\begin{table*}[ht!]
\centering
\scalebox{0.77}{
\begin{tabular}{lcccccccccccc|c}
\toprule
Method 	& \textbf{BN} & \textbf{DE} & \textbf{EN} & \textbf{ES} & \textbf{FA} & \textbf{FR} & \textbf{HI} & \textbf{IT} & \textbf{PT} & \textbf{SV} & \textbf{UK} & \textbf{ZH} & \textbf{AVG.} \\
\midrule
        RaNER w/ one stage & 91.79 & 82.41 & 84.32 & 87.49 & 85.69 & 85.48 & 90.68 & 89.51 & 87.46 & 88.54 & 87.82 & 76.45 & 86.47 \\
\midrule
        RaNER w/ bs 4 & 82.02 & 80.82 & 85.60 & 88.46 & 85.27 & 87.53 & 86.56 & 89.80 & 87.26 & 89.77 & 89.17 & 68.59 & 85.07 \\ 
        RaNER w/ bs 128 & 88.09 & 83.23 & 85.87 & 89.40 & 85.59 & 88.18 & 89.57 & 91.84 & 88.97 & 90.01 & 88.97 & 72.11 & 86.82 \\ 
        RaNER w/ scale up & 90.82 & 86.27 & 85.86 & 89.88 & 86.15 & 88.70 & 90.99 & 91.50 & 89.24 & 90.85 & 88.95 & 75.71 & 87.91 \\ 
\bottomrule
\end{tabular}
}
\caption{The model performance with different training strategies.}
\label{tab:msf}
\end{table*}

\begin{table*}[ht!]
\centering
\scalebox{0.7}{
\begin{tabular}{lcccccccccccc|c}
\toprule
Method 	& \textbf{BN} & \textbf{DE} & \textbf{EN} & \textbf{ES} & \textbf{FA} & \textbf{FR} & \textbf{HI} & \textbf{IT} & \textbf{PT} & \textbf{SV} & \textbf{UK} & \textbf{ZH} & \textbf{AVG.} \\
\midrule
        RaNER & 89.81  & 80.55  & 79.98  & 82.99  & 81.17  & 81.73  & 90.57  & 87.48  & 83.61  & 84.43  & 83.69  & 77.30  & 83.61  \\ 
\midrule
        U-RaNER w/ Pre-infusion & 91.35  & 82.80  & 83.71  & 86.73  & 86.63  & 85.88  & 91.07  & 89.08  & 87.18  & 89.16  & 88.69  & 80.41  & 86.89  \\ 
        U-RaNER w/ Post-infusion & 91.82  & 83.24  & 84.50  & 86.85  & 87.64  & 87.21  & 91.23  & 90.36  & 87.98  & 90.47  & 90.02  & 81.15  & 87.71 \\ 
\bottomrule
\end{tabular}
}
\caption{The model performance with different infusion approaches.}
\label{tab:infuse}
\end{table*}

% case study
\begin{table*}[t]
\centering
\small
\renewcommand\arraystretch{0.7}
\begin{tabular}{lccccc}
    \toprule
        \textbf{Sentence} & \textbf{Span} & \textbf{Gold Tag} & \textbf{BERT-CRF} & \textbf{RaNER} & \textbf{U-RaNER} \\
        \midrule
        \tabincell{l}{pudendal nerve entrapment can \\ occur when the ...}  & 
        \tabincell{c}{pudendal nerve entrapment} & 
        \tabincell{c}{Disease} & 
        \tabincell{c}{-} & 
        \tabincell{c}{Symptom} & 
        \tabincell{c}{Disease} \\ 
        \midrule
        \tabincell{l}{he debuted for gloucestershire \\ in 1887 at the age of ...}  &
        gloucestershire & 
        SportsGRP & 
        SportsGRP & 
        HS & 
        SportsGRP \\
        \midrule
        \tabincell{l}{the main event featured \\ thales leites taking on jesse taylor}  &
        \tabincell{c}{thales leites \\ jesse taylor} & 
        \tabincell{c}{OtherPER \\ OtherPER} & 
        \tabincell{c}{Athlete \\ OtherPER} & 
        \tabincell{c}{Athlete \\ Athlete} & 
        \tabincell{c}{Athlete \\ Athlete} \\
        \bottomrule
    \end{tabular}
\caption{Examples of three NER systems. The entity type {\tt HS} refers to HumanSettlement.}
\label{tab:case_study}
\end{table*}

\subsection{Detailed Procedure for \texttt{TEXT2ENT}}
\label{app:text2ent}
For sparse retrieval, we find the relevant entities from Wikidata which contains millions of entities. As in the \texttt{TEXT2TEXT} strategy, we utilize the description and alias information in the Wikidata and index them with ElasticSearch. We use each sentence in the dataset as the query and retrieve the candidate entity with the BM25 algorithm. In order to find candidate entities as much as possible, we apply an iterative retrieval procedure in which we construct a new query by masking the retrieved entities in the query text from the previous retrieval.

For dense retrieval, we utilize the title information and paragraph information~\footnote{Considering the memory limit of dense retrieval model training, we truncate the paragraph information in wikipedia, and reserve the first 128 tokens for the construction of the knowledge base.} from Wikipedia to construct the knowledge base for dense entity retrieval, then use the input sentence as the query to retrieve its related Top-K entities in the knowledge base. 
The dense retrieval model we use is the widely used Bi-Encoder architecture~\cite{DBLP:conf/iclr/ZhangHS22}.
Different from sparse retrieval, the dense retrieval model is trainable to better perceive the semantic characteristics of the MultiCoNER dataset.  
Therefore, in practice, we first preprocess the train/dev sets of MultiCoNER into the data format for dense retrieval model training. 
Specifically, because the train/dev sets provide the golden entity annotation of the sentence,  we can fuzzy match the span in the sentence with the entity title in our knowledge base to link each span to a specific entity id.
Then we use reconstructed training data to train a dense entity retrieval model with reliable performance, which will be finally applied to the test set to obtain candidate entities for the sentences in the test set.

\subsection{Detailed Procedure for \texttt{ENT2ENT}}
\label{app:ent2ent}
Suppose that we have already retrieved the boundaries of possible or relative entities of a sentence, we want to encode more knowledge about these entities to benefit the prediction of target entities and their types. A good choice is leveraging Wikidata which integrates billions of structural information between millions of entities, such as the alias of entities and the relationships of entity pairs. Therefore, we adopt the following steps to acquire \texttt{ENT2ENT} knowledge to augment the data so as to enhance the entity recognition ability of our model.

\begin{enumerate}
    \item We preprocess Wikidata to construct two dictionaries of each language in this task. One takes each entity name and each alias string of each entity in Wikidata as keys and the index (called ``Qid") of each entity as values. The other takes Qid of each entity as keys and two attributes (called ``subclass of" and ``sub-instance of") content of each entity as values. It is worth mentioning that the values of the two attributes associated with each entity in Wikidata are themselves entities. Therefore, this method is referred to as \textbf{\texttt{ENT2ENT}} retrieval. 
    For the following description, we call the first dictionary \textit{String-to-Qid} and the second dictionary \textit{Qid-to-Types}.
    \item For each language, we retrieve argumentation data according to pre-retrieved entities and the knowledge dictionaries from Step1. Concretely, for each retrieved entity, we first extract the corresponding Qid if it can match one key from the \textit{String-to-Qid} dictionary. Next, if the first operation succeeds, we leverage the Qid to query the \textit{Qid-to-Types} dictionary to get the values of ``subclass of" and ``sub-instance of" as types of the retrieved entity. It is possible that the values of some Qid in the \textit{Qid-to-Types} dictionary of a specific language are NULL. In this situation, we try to get entity types from the \textit{Qid-to-Types} dictionary of English except for processing English itself. 
    \item If we get the language-specific types or English types of some pre-retrieved entities from Step2, we sequentially splice these pre-retrieved entities and their retrieved types after the original sentence. For those pre-retrieved entities without retrieved types, we only splice the pre-retrieved entities.
\end{enumerate}

\subsection{Detailed Procedure for Prompting}
\label{app:prompt}
Following \citep{lai2023chatgpt}, our \texttt{Multi-turn} prompt structure for ChatGPT consists of a task description, a note for output format, and an input sentence.
Since the experiments in \citet{lai2023chatgpt} indicate that English prompts work better than multilingual ones, we use English prompts for all languages.
As shown in Figure \ref{fig:prompt}, the task description part is used to explain the task and list the entity categories, the note part indicates the annotation scheme and output format, and finally we add the input text.
In our experiment, $\{...\}$ is filled by the content in the Appendix \ref{app:taxonomy}.

\texttt{Multi-turn} first performs the task in 6 coarse-grained categories, and later performs finer-grained NER. 
In our experiment, $\{...\}$ is filled by the response of ChatGPT and the content from in the Appendix \ref{app:taxonomy}.

\texttt{Multi-ICL} constructs demonstrations spliced after the note part by randomly selecting examples from the training set. \texttt{xxx} is replaced with the selected example. The corresponding prompts can be found in Figure \ref{fig:prompt_mt}.

\begin{figure}[!ht]
\fcolorbox{bg}{bg}{
\begin{minipage}{18.7em}
    \textbf{Task Description:} %
    You are working as a named entity recognition expert and your task is to label a given text with named entity labels. Your task is to identify and label any named entities present in the text. The named entity labels that you will be using are 33 categories, as shown below \{...\}.\\
    \textbf{Note:} Please use BIO annotation schema to complete this task. Please make sure to label each word of the entity with the appropriate prefix (``B'' for the first word of the entity, ``I'' for any non-initial word of the entity). For words which are not part of any named entity, you should return ``O''. 
    Your output format should be a list of tuples, where each tuple consists of a word from the input text and its corresponding named entity label.\\
    \textbf{Input:} [``from'', ``1995'', ``to'', ``2011'', ``deal'', ``hudson'', ``was'', ``the'', ``magazine's'', ``publisher'', ``.'']\\
    \textbf{Output:}
\end{minipage}
}
\caption{Input prompt for Single-turn.}
\label{fig:prompt}
\end{figure}

\begin{figure}[!ht]
\centering
\fcolorbox{bg}{bg}{
% \scalebox{0.5}{
\scalebox{0.9}{
\begin{minipage}{\linewidth}
% \begin{minipage}{18.7em}
    \textbf{Task Description:} %
    You are working as a named entity recognition expert and your task is to label a given text with named entity labels. Your task is to identify and label any named entities present in the text. The named entity labels that you will be using are PER (person), LOC (location), CW (creative work), GRP (group of people), PROD (product), and MED (medical).\\
    \textbf{Note:} Please use BIO annotation schema to complete this task. Please make sure to label each word of the entity with the appropriate prefix (``B'' for the first word of the entity, ``I'' for any non-initial word of the entity). For words which are not part of any named entity, you should return ``O''. \\
    \textbf{Demonstrations:} Optional. [Input: \texttt{xxx}, Output: \texttt{xxx}]. \\
    \textbf{Input:} [``from'', ``1995'', ``to'', ``2011'', ``deal'', ``hudson'', ``was'', ``the'', ``magazine's'', ``publisher'', ``.'']\\
    \textbf{Output:} \{...\}. \\
    \textbf{Input:} Please complete the above task at a finer granularity based on the fine-grained taxonomy below \{...\}. \\
    \textbf{Output:}
\end{minipage}}
}
\caption{Input prompt for Multi-turn and Multi-ICL.}
\label{fig:prompt_mt}
\end{figure}

\section{More Analysis}

\subsection{Multi-stage Fine-tuning}
\label{app:msf}
We observe that inconsistent training set sizes on different language tracks will lead to degradation of model performance from 86.47\% to 85.07\%. We use increasing batch size and scaling up strategy to address this issue. From the Table \ref{tab:msf}, increasing batch size from 4 to 128 can improve the model performance from 85.07\% to 86.82\%.
Furthermore, scaling up the training data size on \texttt{BN}, \texttt{DE}, \texttt{HI} and \texttt{ZH} can also result in a gain of +1.09\%

\subsection{Two Infusion Approaches}
\label{app:infuse}
In the section \cref{4.2}, we propose two infusion methods (Pre-Infusion and Post-Infusion) to make more context visible to the model. 
Here, we make a quantitative comparison of their effects on model performance.
As shown in the Table \ref{tab:infuse},
we observe that the post-infusion method is superior to the pre-infusion method in all language track. 
We attribute this to the fact that the pre-infusion method only considers the anchor information and ignores other contextual information, while the post-infusion method uses more contextual knowledge and achieves better performance.

\begin{table}[t!]
\centering
\setlength{\tabcolsep}{1.9mm}
\scalebox{0.75}{
\renewcommand\arraystretch{1.1}
\begin{tabular}{lccccc|c}
\toprule
        \textbf{Metric} & \textbf{BN} & \textbf{DE} & \textbf{PT} & \textbf{SV} & \textbf{ZH} & \textbf{AVG.} \\ 
\midrule
        Recall-Sparse & 79.32  & 71.09  & 98.22  & 98.20 & 37.76  & 76.92  \\ 
        Recall-Dense & 93.26 & 85.18 & 87.84 & 89.19 & 79.80 & \textbf{85.25} \\ 
\midrule
        F1-Baseline & 86.98  & 85.46  & 76.54  & 78.48  & 75.11  & 80.51  \\ 
        F1-Sparse & 89.81  & 90.57  & 83.61  & 84.43  & 77.30  & \textbf{85.14}  \\ 
        F1-Dense & 88.45  & 89.83  & 77.23  & 80.54  & 78.00  & 82.81 \\ 
\bottomrule
\end{tabular}
}
\caption{Comparison of retrieval performance and impact on NER between the sparse and dense \texttt{TEXT2ENT} strategies on the dev set.}
\label{tab:different_retrieval_methods}
\end{table}

\subsection{Different Retrieval Methods}
\label{app:retrieval}
To deeply analyze the effectiveness of the two \texttt{TEXT2ENT} retrieval strategies we design, we compare their retrieval performance (i.e., Recall@50) and the enhanced NER performance (i.e., F1) based on their respective retrieval results.
From Table~\ref{tab:different_retrieval_methods}, we find that the retrieval performance of sparse retrieval does not seem to be worse than dense, and its recall is higher than dense retrieval for both PT and SV languages. In addition, for the BN and DE languages, although their recall results of sparse retrieval are lower than those of dense retrieval, their final performance of NER is higher than that of dense retrieval. 
We think this is mainly due to the different retrieval sources of the two retrieval strategies. Our sparse strategy is retrieved from Wikidata, while the dense strategy is retrieved from Wikipedia. The retrieval quality of Wikipedia is easily disturbed by the existence of entity alias. In addition, because the dense retrieval requires us to train the model, we actually truncate the paragraph information in Wikipedia for model training and retrieval, so the information that can be used for dense retrieval is also limited. 
However, from the ZH language, we know that the robustness of the dense retrieval strategy for different languages is better than the sparse retrieval strategy. Therefore, when dealing with retrieval in different languages, we can flexibly choose different strategies based on the quality of the retrieval resources in the corresponding language to obtain better performance.

\subsection{Case Study}
\label{app:case}

Table ~\ref{tab:case_study} provides a closer examination of the predicted results of BERT-CRF, RaNER, and U-RaNER respectively. We selected three cases from the English language dev data to analyze in detail.

In the first case, fine-grained NER necessitates comprehensive information to accurately classify long-tail entity spans. By utilizing knowledge from multiple sources, U-RaNER successfully predicts "pudendal nerve entrapment" in the first case.

In the second case, RaNER's typical ambiguity problem is evident, where the context retrieved from merely Wikipedia source lacks pertinent information about the target entity "gloucestershire" which could refer to either a county or a sports club.

However, in the third case, the retrieval-based systems wrongly predict "theles leites" and "jesse taylor" as "Athlete" due to retrieved knowledge indicating that they are both mixed martial arts fighters. This demonstrates that the use of retrieved information can sometimes be misleading and even harmful.

\end{document}